\documentclass{article}

\usepackage{arxiv}

\usepackage[utf8]{inputenc} 
\usepackage[T1]{fontenc}    
\usepackage{hyperref}       
\usepackage{url}            
\usepackage{booktabs}
\usepackage{tikz}
\usepackage{amsmath}
\usepackage{amsfonts}
\usepackage{mathtools}
\usepackage{natbib}
\usepackage{xcolor}
\usepackage[toc, page]{appendix}
\usetikzlibrary{external}
\newcommand\given{\,|\,}

\newcommand{\zero}{\mathbf 0}

\newcommand{\KL}{\textsc{KL}}
\newcommand{\Yml}{\phi_\ell^m}
\newcommand{\Ymlp}{\phi_{\ell^\prime}^{m^\prime}}
\newcommand{\uml}{u_\ell^m}
\newcommand{\fml}{\hat{f}_{\ell,m}}
\newcommand{\Nld}{N(\ell,d)}
\newcommand{\diff}{\mathop{}\!\mathrm{d}}

\title{Sparse Gaussian Processes with Spherical Harmonic Features Revisited}
\author{%
  Stefanos Eleftheriadis\thanks{Amazon.com}%
  \And Dominic Richards\footnotemark[1]%
  \And James Hensman\thanks{Work done while at Amazon; currently with Microsoft Research}%
  }

\begin{document}
\maketitle

\begin{abstract}
We revisit the Gaussian process model with spherical harmonic features and study connections between the associated RKHS, its eigenstructure, and deep models.
Based on this, we introduce a new class of kernels which correspond to deep models of continuous depth. 
In our formulation, depth can be estimated as a kernel hyper-parameter by optimizing the evidence lower bound. 
Further, we introduce sparseness in the eigenbasis by variational learning of the spherical harmonic phases.
This enables scaling to larger input dimensions than previously, while also allowing for learning of high frequency variations.
We validate our approach on machine learning benchmark datasets.
\end{abstract}

\section{Introduction}
Deep neural networks (DNNs) have become a go-to methodology for modeling functional relationships from data, $y = f(x) +
\epsilon$. The ability to
train massive networks on structured data has led to impressive results on language modeling and computer vision tasks.
Meanwhile, scientists in other fields seek reliable function approximators with statistical uncertainty quantification,
and in these applications Gaussian process (GPs) models have become the standard. Is there a way to get the best of both? Can
we take some of the model-structure and scalability that have made DNNs so popular, and embed it into a Gaussian process
model? 

Recent works \citep{dutordoir2020sparse, dutordoir2021deep} have made use of spherical harmonic analysis to
draw connections between Gaussian process methods and deep networks. Those authors introduced a variational inference
method where spherical harmonics form the basis of an approximation to the Gaussian process posterior.
Since the
spherical harmonics form a Karhunen-Loeve expansion of the Gaussian process, several matrix computations are reduced in
complexity due to analytic diagonalization. This methodology forms the basis of the current manuscript and we give an
introduction in section \ref{sec:variational_intro}. 

Meanwhile, other authors have explored the connections between the limits of very large DNNs and Gaussian
processes. In his seminal thesis, \citet{neal1996priors} showed that a single-layer neural network with a Gaussian prior on
the weights becomes a Gaussian process as the size of the network grows to infinity. \citet{cho2009kernel} introduced deep
kernels by allegory with deep networks, and \citet{matthews2018gaussian} showed that these kernels do indeed correspond to
multi-layer neural networks with Gaussian process priors. They also showed that in practical settings, the Gaussian
process limit arises rather quickly, with networks of width of order 100 behaving as Gaussian processes. 

Further work has demonstrated that Gaussian process behavior of deep networks remains when the network is trained
using gradient descent. \citet{jacot2018neural} introduced the neural tangent kernel (NTK), which describes how a
trained neural network exhibits Gaussian process behavior in the large-width regime. \citet{yang2020tensor} devised a
systematic method to compute such a kernel corresponding to a large number of neural network architectures, and
\citep{garriga2018deep} examined the case of convolutional residual networks. A common theme in these works is that
whilst Gaussian-process equivalents to large networks exist, they are often expensive to compute. For example in
\citet{bietti2021approximation}, constructing the kernel matrix for an image-recognition task took 10 hours on a
1000-core machine. 

This work brings together the computational method from \citet{dutordoir2020sparse} with some of the advances in understanding the (kernels of) Gaussian
processes connection to large DNNs. We first review some connections between RKHS eigenstructures and deep model
structures; we clarify how the corresponding RKHS gives rise to polynomial kernels with spherical harmonics as the orthogonal basis;
we introduce kernels of continuous depth so that depth may be estimated as a kernel hyper-parameter;
and we introduce variational learning of spherical-harmonic phases, which enables scaling to larger input dimensions than
previously.

We demonstrate our combined methods on machine learning benchmark datasets.

\section{Deep learning with kernels on the sphere}
By definition, any function that can be factorised into a radial and an angular component,
must be acting on a hyper-sphere, i.e., it is a spherical function. For instance, let us take the typical Relu function defined as $\sigma_{\textrm{relu}}(\mathbf{x}^\top\mathbf{w}) = \max(0, \mathbf{x}^\top\mathbf{w})$, for some $\mathbf{w} \sim \mathcal N(\zero, \mathbf{I})$, $\mathbf{x} \in \mathbb R^d$. If we project the vectors onto the unit sphere $\mathbb S^{d-1}$ by normalising them, we can easily see that the Relu function is indeed spherical:
\begin{equation}
    \sigma_{\textrm{relu}}(\mathbf{x}^\top\mathbf{w}) = \|\mathbf{x}\| \|\mathbf{w}\|\max(0, \frac{\mathbf{x}^\top\mathbf{w}}{\|\mathbf{x}\| \|\mathbf{w}\|}) = \underbrace{\|\mathbf{x}\| \|\mathbf{w}\|}_{\textrm{radial}} \underbrace{\sigma_{\textrm{relu}}(\frac{\mathbf{x}^\top\mathbf{w}}{\|\mathbf{x}\| \|\mathbf{w}\|})}_{\textrm{angular}}\,.
\end{equation}

Based on this observation, \citet{cho2009kernel} studied the limit of infinite wide fully connected neural networks when the activation is a Relu function.
They found that the equivalent kernel takes the form of:
\begin{align}\label{eq:k_relu}
    k(\mathbf{x}, \mathbf{x}^\prime) = \mathbb E_\mathbf{w} \left[\sigma_{\textrm{relu}}(\mathbf{w}^\top\mathbf{x})\sigma_{\textrm{relu}}(\mathbf{w}^\top\mathbf{x}^\prime)\right]
    = \underbrace{\|\mathbf{x}\| \|\mathbf{x}^\prime\|}_{\textrm{radial}}
    \underbrace{\frac{1}{\pi}\left(t\left(\pi - \textrm{arccos}(t)\right) + \sqrt{1 - t^2}\right)}_{\textrm{angular}, \kappa(t)}\,,
\end{align}
with $t = \frac{\mathbf{x}^\top\mathbf{x}^\prime}{\|\mathbf{x}\| \|\mathbf{x}^\prime\|}$ and $\kappa(t)$ the {\em shape function} of the kernel.
We see that the above kernel is a bi-zonal function (zonal in either $\mathbf{x}$ or $\mathbf{x}'$).
This is an improtant observation, since zonal functions enjoy a particular relation with spherical harmonics, as we shall see in the next section.

\paragraph{Equivalent kernel of a deep network.} The above result can be extended to derive the equivalent kernel of a deep fully connected network with more than two layers.
Allowing the layers to be wide enough, i.e., infinite width we end up with the not
surprising result of:
\begin{equation}
    \kappa^L(t) \coloneqq \underbrace{\kappa \circ \cdots \circ \kappa}_{L-1 \textrm{ times}}(t).
\end{equation}
All we left to do is to rescale the shape function by multiplying with the appropriate radii so we have the final form of the equivalent kernel. Note that a good practice is to normalise $\kappa(1) = 1$, so that we also have $\kappa^L(1) = 1$.

\section{Introduction to spherical harmonics}
Spherical harmonics are functions defined on the surface of a sphere.
They arise as the solution of the angular part of Laplace's equation when expressed in spherical coordinates.
Given a point on the unit hyper-sphere, $\mathbf{x} \in \mathbb S^{d-1}$,
we write $\Yml(\mathbf{x})$ to denote the spherical harmonic of order (or frequency) $\ell$ and orientation (or phase) $m$, with $\ell \ge 0$ and $\lvert m\rvert \le \ell$.
Spherical harmonics are understood to be the generalization of a Fourier series to the sphere: the order $\ell$ is equivalently a frequency, and the orientations $m$ are phases. For a 2-sphere, the harmonics correspond precisely to sines and cosines of frequency $2\pi\ell$, and there are exactly two phases for any frequency (i.e. the sine and cosine part of the Fourier series). 

An important property of the spherical harmonics is that they form a complete, orthonormal basis on the sphere $\mathbb S^{d-1}$, which is embedded in the $d$-dimensional space $\mathbb R^d$. Therefore, they satisfy the property:
\begin{equation}
    \int_{\mathbb S^{d-1}} \Yml(\mathbf{x})\Ymlp(\mathbf{x}) \diff\Omega
     = \delta_{\ell\ell^\prime}\delta_{mm^\prime}\,,
\end{equation}
where $\delta$ is Kronecker's delta and $\Omega$ the surface of the $\mathbb S^{d-1}$ sphere. For a specific dimension $d\geq 3$ and order $\ell$ there exist
\begin{equation}
    \Nld = \frac{2\ell + d - 2}{\ell} \begin{pmatrix}
        \ell + d - 3\\
        d - 1
\end{pmatrix}
\end{equation}
linearly independent harmonics.

Any zonal function can be written as the linear combination of the spherical harmonics:
\begin{equation}\label{eq:zonal_f}
    f(\mathbf x^\top \mathbf{z}) = \sum_{\ell=0}^{\infty} \sum_{m=1}^{\Nld} \fml \Yml(\mathbf x)\,,
\end{equation}
where the coefficients $\fml$ are given by the Funk-Hecke formula:
\begin{equation}
    \int_{\mathbb S^{d-1}} f(\mathbf{x}^\top \mathbf{z})\,\Yml(\mathbf x) \diff\Omega = \lambda_\ell\Yml(\mathbf z) \eqqcolon \fml\,.
    \label{eq:eigval_f}
\end{equation}
We identify the terms $\fml$ as the Fourier coefficients of the function $f$, associated with the eigenfunctions $\Yml$, i.e., the spherical harmonics. The terms $\lambda_\ell$ are the eigenvalues, which as we will see in a next section do not depend on the orientation $m$.

 Another property of the spherical harmonics that will be proven useful in our analysis comes from the {\em addition} theorem. This states that for the spherical harmonics of degree $\ell$ in dimension $d$ the following holds:
 \begin{equation}\label{eq:addition_thm}
     \sum_{m=1}^{\Nld} \Yml(\mathbf{x})\Yml(\mathbf{x}^\prime) = \frac{\ell + \alpha}{\alpha}C_\ell^{(\alpha)}(\mathbf{x}^\top\mathbf{x}^\prime)\,,
\end{equation}
where $\alpha = \frac{d-2}{2}$ and $C_\ell^{(\alpha)}$ is the Gegenbauer polynomial of order $\ell$.

\subsection{Linear models on the spherical harmonic basis}
Spherical harmonics can be used as basis function in linear model in the same way as any other basis function. Since linear-Gaussian models can be written as Gaussian processes, it should be clear that linear-Gaussian models with spherical harmonic basis lead to Gaussian processes with spherical kernels. To illustrate, consider a linear model for a single frequency $\ell$ of the form $g(\mathbf{x}) = \sum_m w_m\Yml(\mathbf{x})$, with $\mathbf{x} \in \mathbb S^{d-1}$ and $w_m \sim \mathcal N(0, \lambda_\ell)$, where we explicitly pick $\lambda_\ell$ as the variance to make the connection obvious.
Taking the product of the function at two points $\mathbf x$ and $\mathbf x^\prime$ and integrating over $w_m$ we have:
\begin{align}\label{eq:dot_kernel}
    \mathbb E_{w_m}\left[g(\mathbf x)g(\mathbf x^\prime)\right] =
    \sum_m^{\Nld} \mathbb E_{w_m}\left[w_m^2\right] \Yml(\mathbf{x})\Yml(\mathbf{x}^\prime) = 
    \lambda_\ell \frac{\ell + \alpha}{\alpha} C_\ell^{\alpha}(\mathbf{x}^\top\mathbf{x}^\prime)\,,
\end{align}
where in the first equality we used the independence between $w_m$ and in the second equality the addition theorem \eqref{eq:addition_thm}.

We recognize the right-hand-side of Eq.~\eqref{eq:dot_kernel} as a kernel.
With a closer look, we see that this kernel is a bi-zonal function, containing only the $\ell$th frequency, and can be written in the form of Eq.~\eqref{eq:zonal_f}. Now plugging Eq.~\eqref{eq:eigval_f} into Eq.~\eqref{eq:zonal_f} and applying the addition theorem from Eq.~\eqref{eq:addition_thm}, recovers the exact same kernel.
This gives rise to the reproducing property of the kernel space.
We continue our analysis by properly defining spherical kernels via the corresponding RKHS.


\subsection{From spherical functions to spherical kernels}
Following Mercer's theorem, the RKHS $\mathcal H$ associated to a zonal kernel is given by:
\begin{equation}
    \mathcal H = \{f: \sum_{\ell\ge 0} \sum_{m=1}^{\Nld} \fml\Yml(\cdot)\quad \textrm{s.t.}\quad\lvert\lvert f \rvert\rvert^2_\mathcal{H} \coloneqq \sum_{\ell\ge0, \lambda_\ell \neq 0}\sum_{m=1}^{\Nld}\frac{\fml^2}{\lambda_\ell} < \infty\}\,,
\end{equation}
where $\lambda_\ell$ is the eigenvalue of the kernel corresponding to the $\ell$th frequency.

To compute the eigenvalues $\lambda_\ell$ we first observe that for a given point $\mathbf{x^\prime} \in \mathbb S^{d-1}$, the kernel $k(\mathbf{x}, \mathbf{x}^\prime) : \mathbb S^{d-1} \rightarrow \mathbb R$ is a spherical function. As such, it can be factorised in a radial (i.e., the scale) and an angular (i.e., the associated shape function) component. Further, it can be written as a linear combination of the spherical harmonics
\begin{equation}\label{eq:kernel}
    k(\mathbf{x}, \mathbf{x}^\prime) \coloneqq r(\mathbf x, \mathbf x^\prime) \kappa(\mathbf{x}^\top\mathbf{x}^\prime) = r(\mathbf x, \mathbf x^\prime)\sum_{\ell=0}^{\infty} \sum_{m=0}^{\Nld} \lambda_\ell \Yml(\mathbf{x}) \Yml(\mathbf{x}^\prime)\,,
\end{equation}
where $\kappa(\cdot)$ is the shape function (i.e., the angular component) of the kernel.
Similarly to Eq.~\eqref{eq:eigval_f} we express $\lambda_\ell$
\begin{equation}
    \lambda_\ell \Yml(\mathbf x) =  
    \int_{\mathbb S^{d-1}} \kappa(\mathbf{x}^\top\mathbf{x}^\prime)\,\Yml(\mathbf{x}^\prime) \diff\Omega\,,
    \label{eq:eigval_k}
\end{equation}
and the $(d-1)$dimensional integral can be transformed to a one dimensional over the shape function via the Funk-Hecke formula to give us the eigenvalues:
\begin{equation}
   \lambda_\ell =  
    \frac{\omega_d}{C_\ell^{(\alpha)}(1)}\int_{-1}^1 \kappa(t) C_\ell^{(\alpha)}(t)(1 - t^2)^\frac{d-3}{2} \diff t\,,
    \label{eq:eigval_FH}
\end{equation}
$\omega_d$ is the surface of the sphere.

Combining Eqs.~\eqref{eq:kernel} and~\eqref{eq:addition_thm} enable us to write any spherical kernel as a polynomial expansion:
\begin{align}\label{eq:k_expanse}
    k(\mathbf{x}, \mathbf{x}^\prime) = r(\mathbf x, \mathbf x^\prime)
    \sum_{\ell=0}^{\infty} \frac{\ell + \alpha}{\alpha} \lambda_\ell C_\ell^{(\alpha)}(\mathbf{x}^\top\mathbf{x}^\prime)\,.
\end{align}

\section{Continuous depth kernels}
In the previous section we have seen how to construct equivalent kernels for fully connected deep neural networks, by simple composition of the shape function.
Here we introduce our approach on defining spherical kernels with continuous depth.

We begin by inspecting Eq.~\eqref{eq:k_relu} for the case of $L$ compositions of the shape function, $\kappa^L(\cdot)$.
If we are able to compute the eigenvalues of the deep Relu kernel, or any other kernel with a valid shape function, then we can write it as a polynomial expansion, similarly to Eq.~\eqref{eq:k_expanse}.
The behaviour of the eigenvalues for the spherical kernels has been the subject of interest in many studies recently~\citep{bietti2021deep,belfer2021spectral}. One key result is that the eigenvalues decay polynomially as we move to higher frequencies.
Moreover,~\citet{bietti2021deep} have shown that for different shape functions, the decay rate can depend or the depth.

Motivated from the above findings and looking again at Eq.~\eqref{eq:k_expanse} we can define a spherical kernel without having access to a specific shape function and without needing to compute the integral from Eq.~\eqref{eq:eigval_FH}, which obviously depends on the depth through the composed shape function.

We propose to use a kernel of the form:
\begin{align}\label{eq:k_poly}
    k(\mathbf{x}, \mathbf{x}^\prime) = r(\mathbf x, \mathbf x^\prime)
    \sum_{\ell=0}^{\infty} \frac{\ell + \alpha}{\alpha} {\ell}^{-\beta} C_\ell^{(\alpha)}(\mathbf{x}^\top\mathbf{x}^\prime)\,,
\end{align}
where we explicitly model the eigenvalues via the polynomial $\ell^{-\beta}$ and, the order of the polynomial $\beta > 0$ is a kernel hyper-parameter.
There are two important observations we need to state. First, the proposed kernel does not correspond directly to a known function; instead, it can be seen as a random spherical harmonic feature expansion, in analogy to the random Fourier features. Second, by learning the $\beta$ hyper-parameter we model the effect of depth in a continuous way; lower values for $\beta$ correspond to deeper kernels.

\begin{figure}[ht]
\includegraphics[scale=0.4, clip]{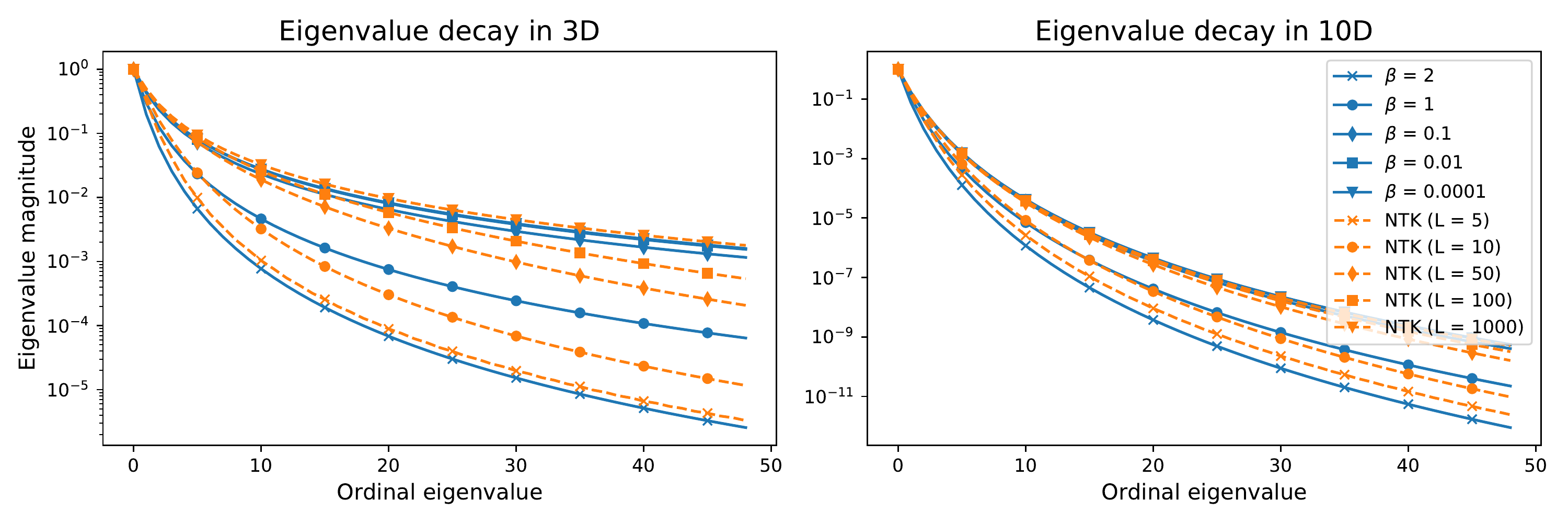}
\centering
    \caption{Eigenvalue comparison between the proposed kernel with continuous depth vs the NTK kernel with varying depth levels, for 3 dimensional problems (left) and 10 dimensional problems (right). Each eigenvalue is depicted relative to the corresponding eigenvalue of the first non-constant frequency.}
\label{fig:eigvals}
\end{figure}

\paragraph{Empirical study of the eigenvalue decay.} In Fig.~\ref{fig:eigvals} we compare the eigenvalue decay between the two approaches: (i) the deep NTK kernel~\citep{jacot2018neural} via compositions of the shape function; and (ii) the proposed polynomial expansion approach with continuous depth via the hyper-parameter $\beta$.
The first thing to notice is that the dimension of the problem plays an important role. Counter-intuitively, in higher dimensions (right panel), the high frequencies vanish rapidly, as even after only $10$ frequencies the decay is more than three orders of magnitude.
Further, by adding more depth the decay rate becomes slower in both low and high dimensions and the high frequencies become more relevant. The effect is especially pronounced in the low dimensional problems.
Finally, notice how the proposed kernel with continuous values for $\beta$ mimics the behaviour of NTK with increasing number of shape function compositions, i.e., depth.

\section{Spherical harmonics as features for GPs}
Here we leverage our knowledge on the spherical kernels to build efficient Gaussian process models on the associated RKHS. These models mimic the behavior of deep fully connected neural networks.

\subsection{Inter-domain GPs with spherical harmonics}\label{sec:variational_intro}
In their work~\citet{dutordoir2020sparse} have proposed a variational approach based on spherical harmonics to learn an approximation to the GP posterior.
To do so they introduced an inter-domain approach where the inducing features are the spherical harmonics. This results in diagonal covariance structure for the kernel via the Mercer's theorem and also allows to learn features with global structure compared to the local information of the traditional inducing points.
More specifically, they define inducing variables $\uml$ as the inner product between the GP function $f$ and the spherical harmonics:
\begin{equation}
    \uml = \langle f, \Yml \rangle_\mathcal{H}.
\end{equation}
Then, they use the reproducing property to compute the covariance between the function and the inducing features, i.e., $\left[k_{fu}(\mathbf x)\right]_{\ell,m}$ as:
\begin{equation}\label{eq:kuf}
    \textrm{cov}\left[f(\mathbf x), \uml\right] = \mathbb E\left[f(\mathbf{x})\uml\right] = \langle k(\mathbf{x}, \cdot), \Yml \rangle_\mathcal{H} = \Yml(\mathbf{x})\,.
\end{equation}
Similarly for the covariance between the inducing features they obtain:
\begin{align}\label{eq:kuu}
    \textrm{cov}\left[\uml, u_{\ell^\prime}^{m^\prime}\right] =
    \mathbb E\left[\uml u_{\ell^\prime}^{m^\prime}\right] =
    \langle \langle k(\cdot, \cdot), \Yml \rangle_\mathcal{H}, \Ymlp \rangle_\mathcal{H} = 
    \langle\Yml, \Ymlp \rangle_\mathcal{H} = 
    \frac{\delta_{\ell\ell^\prime}\delta_{mm^\prime}}{\lambda_\ell}\,,
\end{align}
which admits a diagonal structure for the kernel matrix $\mathbf{K}_{uu}$.

Plugging Eqs.~\eqref{eq:kuf},\eqref{eq:kuu} into the variational posterior from~\citep{hensman2013gaussian} leads to the approximation
\begin{align}
    q(f) = \mathcal{GP}\left(
    \mathbf{\Phi}^\top(\cdot)\mathbf{m},
    k(\cdot,\cdot) + \mathbf{\Phi}^\top(\cdot)(\mathbf{S} - \mathbf{K}_{uu})\mathbf{\Phi}(\cdot)
    \right)\,,
\end{align}
where $\mathbf{\Phi}(\cdot) = \{\lambda_\ell \Yml(\cdot)\}_{\ell,m}$ and $\mathbf{m}, \mathbf{S}$ are the mean and variance, respectively, of the variational distribution $q(\mathbf{u}) = \mathcal{N}(\mathbf m, \mathbf S)$.

To learn the model, one needs to optimise the evidence lower bound (ELBO) wrt the variational parameters and the kernel hyper-parameters
\begin{align}
    \textrm{ELBO} = \sum_{i=1}^N \mathbb E_{q(f)}\left[\log p(y_i\given f(\mathbf{x}_i)) \right]
     - \KL\left[q(\mathbf{u})\,||\,p(\mathbf{u})\right]\,,
\end{align}
where $\mathbf{y}$ is the output of the function we are trying to model, $p(y_i | f(\mathbf{x}_i))$ is the likelihood of choice and, $p(\mathbf{u}) = \mathcal{N}(\zero, \mathbf{K}_{uu})$ is the sparse GP prior. 

\subsection{Sparse features with phase truncation}
From a practical perspective, working with the method from~\citet{dutordoir2020sparse} requires to pick a truncation level $\hat{\ell}$ for the order/frequency of the spherical harmonics.
Then a set of points $\mathbf{V} = \{ \mathbf{v}_{\ell,m}\}_{\ell, m}, \mathbf{v}_{\ell,m}\in\mathbb S^{d-1}$, is chosen on the sphere via a Gram-Schmidt orthogonalisation on $\mathbf{V}_\ell^\top \mathbf{V}_\ell$, so that the points $\mathbf{V}_\ell = \{\mathbf{v}_{\ell,m} \}_{m=1}^{\Nld}$ are maximally separated and form a complete fundamental set. Then, $\{C_\ell^{(\alpha)}(\mathbf{V}_\ell^\top \mathbf{V}_\ell) \}_{\ell=0}^{\hat{\ell}}$ corresponds to the full set of spherical harmonic features up to order $\hat{\ell}$.
The variables $\mathbf{V}$ play the role of the inducing inputs and are kept fixed throughout optimization, as they are already optimally placed.

Although efficient, this approach has two limitations. First, the number of inducing points $M$, which is the total number of phases across all the chosen frequencies, scales exponentially with the number of dimensions. Second, most of the high-frequency components are explicitly ignored due to the truncation at a lower frequency $\hat{\ell}$.

To alleviate this, we propose to introduce an extra truncation $\hat{m}$, this time at the phase level of the spherical harmonics.
So instead of using all the harmonics within each frequency, we settle for a smaller number of basis functions. 
This practically allows us to truncate the frequencies at a much higher order $\hat{\ell}$, which results in features that capture more high frequency characteristics of the function.

A direct consequence of our proposed approach is that the frequencies that have been truncated in phase do not constitute a set of spherical harmonics any more. We ensure, however, that they remain orthogonal polynomials by explicitly orthogonalizing the corresponding $\mathbf{V}_\ell$ within each truncated frequency. Furthermore, we now have the option to optimize the phases of the truncated frequencies, as the $\mathbf{v}_{\ell,m}$ are variational parameters in our ELBO.

We call the features learned by this two-way truncation as {\em sparse} spherical harmonic features.

\section{Experiments}
Here we evaluate the proposed methodology in the context of sparse GPs using the continuous depth kernel and the sparse spherical harmonics as inducing features.

\subsection{Experiment on SUSY classification}
We first demonstrate the effectiveness of the proposed approach on a large-scale classification problem.
To do so, we use the SUSY dataset to classify whether we can detect from the result of a simulation if super-symmetric (SUSY) particles have been produced or not.
We follow a similar experimental setup as~\citet{dutordoir2020sparse} and we use the last 10\% out of the 5 million records in the dataset to test our model.
The inputs to the model consist of eight kinematic properties measured by the
particle detectors in the accelerator.
To train the models we used $7$ frequencies and with a truncation at $100$ phases. This results to approximately $500$ inducing features.
In Table~\ref{tab:susy}, we report the results in terms of the AuC score and we compare against
a custom 5 layer neural network architecture from~\citep{baldi2014searching}.
We see that our single layer GP performs similarly to a DNN without any hassle on picking an architecture, or deciding on the size of the depth.
For reference we have also included the performance of the classic spherical harmonics features for sparse GPs from~\citep{dutordoir2020sparse}, which we have trained with $5$ frequencies with no phase truncation (results again in approximately $500$ inducing features).
It is important to note that the GP with the proposed sparse spherical features have reached the performance of~\citep{dutordoir2020sparse} in half the iterations and then the AuC continued to improve.

\begin{table*}[tbh]
    \caption{Performance comparison on the SUSY dataset.}
  \label{tab:susy}
  \centering
  \begin{tabular}{lc}
    \toprule
    Method  & AuC\\
    \midrule
      ours & $0.870$\\
      \citet{dutordoir2020sparse} & $0.864$\\
      \citet{baldi2014searching} & $0.867$\\
    \bottomrule
  \end{tabular}
\end{table*}

\subsection{Experiment on UCI regression benchmarks}
\begin{figure}[ht]
\includegraphics[scale=0.4, clip]{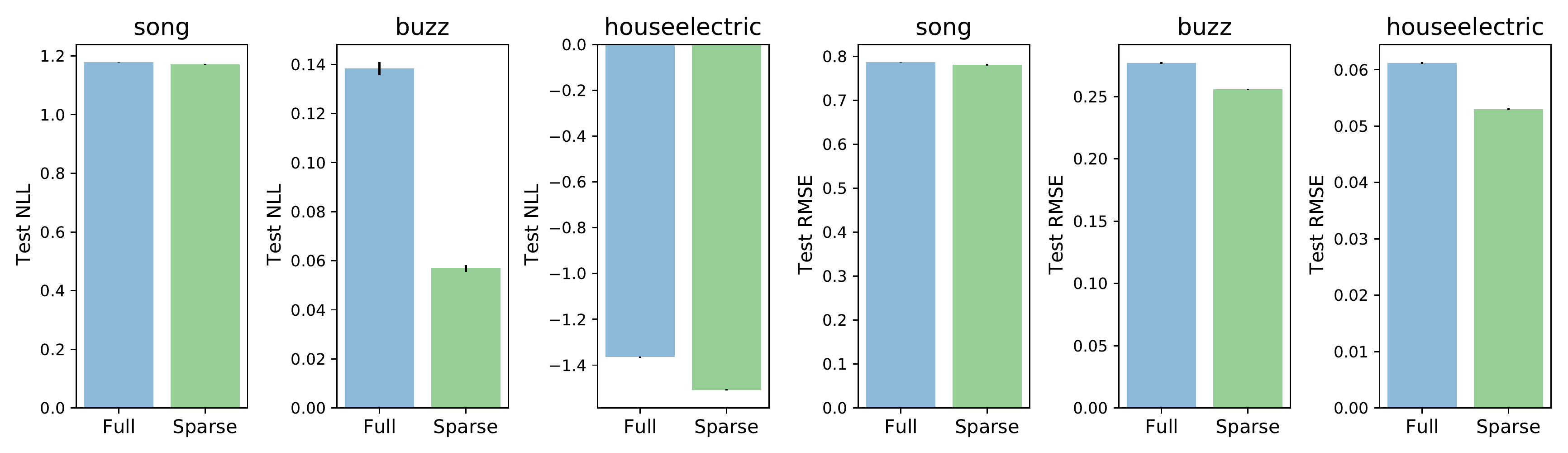}
\centering
    \caption{Test negative log-likelihood and root mean squared error on regression benchmarks. The sparse spherical harmonic features outperform the full harmonics most of the time.}
\label{fig:uci}
\end{figure}

We continue our evaluations on the \textit{song}~\citep{bertin2011million}, \textit{buzz} and \textit{houseelectric} datasets from the UCI corpus~\citep{dua2019}.
We follow the same experimental setup as in~\citep{sun2021scalable} and randomly choose $20\%$ of the data points as test set and repeat across $3$ splits.
We compare the sparse spherical harmonic features with $15$ frequencies and phase truncation of $100$ ($\sim1500$ inducing features), to the model with all the harmonics with $5$ frequencies and no phase truncation ($\sim1400$--$1700$ inducing features). Both models are trained with the polynomial decay kernel. Fig.~\ref{fig:uci} summarizes the results in terms of the test negative log-likelihood (NLL) and the root mean squared error (RMSE). It is worth noting that the sparse spherical features not only outperform the classic spherical harmonics but they also achieve superior performance compared to~\citep{sun2021scalable} which needs $16K$ inducing features.

\section{Conclusions}
In this work we revisited the prior work on sparse Gaussian processes with spherical harmonics 
to solidify the understanding of the connection between deep models and spherical functions.
Specifically, we introduced a new kernel which corresponds to deep models of continuous depth and we further proposed to variationally learn the phases of spherical harmonic features, which results in
a more efficient set of global descriptors with high frequency components.
Our experimental evaluations on standard machine learning benchmarks verify the efficacy of the proposed approach.

\bibliographystyle{apalike}
\bibliography{refs}

\end{document}